\documentclass{article}
\usepackage{aaai}
\usepackage{times}
\usepackage{helvet}
\usepackage{courier}
\usepackage{multicol}
\usepackage{subfig}
\usepackage{stfloats}
\usepackage{hyperref}
\usepackage{url}

\usepackage {graphicx}
\usepackage{array}
\usepackage{amssymb}
\usepackage{dsfont}
\title{Three Tiers Neighborhood Graph and Multi-graph Fusion Ranking for
Multi-feature Image Retrieval: A Manifold Aspect}
\author{Shenglan Liu, Muxin Sun, Lin Feng, Yang Liu, Jun Wu\\
\emph{Faculty of Electronic Information and Electrical Engineering, }\\\emph{Dalian University of Technology, Dalian, Liaoning, 116024 China}}

\begin{document}
\maketitle

\begin{abstract}
Abstract: Single feature is inefficient to describe content of an image,
which is a shortcoming in traditional image retrieval task. We know that one
image can be described by different features. Multi-feature fusion ranking
can be utilized to improve the ranking list of query. In this paper, we first analyze graph structure and multi-feature fusion re-ranking from manifold aspect. Then, Three Tiers Neighborhood Graph (TTNG) is constructed to re-rank the original ranking list by single feature and to enhance precision of single feature. Furthermore, we propose Multi-graph Fusion Ranking (MFR) for multi-feature
ranking, which considers the
correlation of all images in multiple neighborhood graphs. Evaluations are
conducted on UK-bench, Corel-1K, Corel-10K and Cifar-10 benchmark datasets.
The experimental results show that our TTNG and MFR outperform than
other state-of-the-art methods. For example, we achieve competitive results
N-S score 3.91 and precision 65.00{\%} on UK-bench and Corel-10K datasets
respectively.
\end{abstract}


\section{Introduction}

Image ranking has made a number of significant achievements in image
retrieval tasks. Ranking methods have attracted increasing attention to image retrieval. In most cases, we usually utilize 1-norm to measure the similarity for statistical histogram of image feature in ranking
stage. This direct similarity metric ranking results can be regarded
as K-nearest neighborhood (KNN) of query (in re-ranking methods known as a Candidate KNN Set
(CKNNS)). However, the K-nearest neighborhood of query is independent to each other, that is,
there is no connection between the images of the retrieval results. In
general, we assume: \textit{the KNN of query (including query) are similar images and should be related in image retrieval}. This relationship is conducive to the elimination of
outlier in the CKNNS, which is conducive to enhance the results of image
retrieval. Image re-ranking methods can be developed by
CKNNS of query.

 This paper focuses on selection of CKNNS and re-ranking for image retrieval by manifold way. Most of previous research only consider the similarity or using graph method to enhance retrieval results. However, image manifold in real word is always complex, which may not be suitable for image retrieval. Note that \emph{images that are
closer\footnote{ We use Jaccard coefficient to measure the similarity of
images.} to the query image may not have higher correlation with the query
image} which is a serious shortcoming of image manifold for image retrieval task. An example is shown in Fig
1.

\begin{figure}[htbp]
\centerline{\includegraphics[width=3.5in]{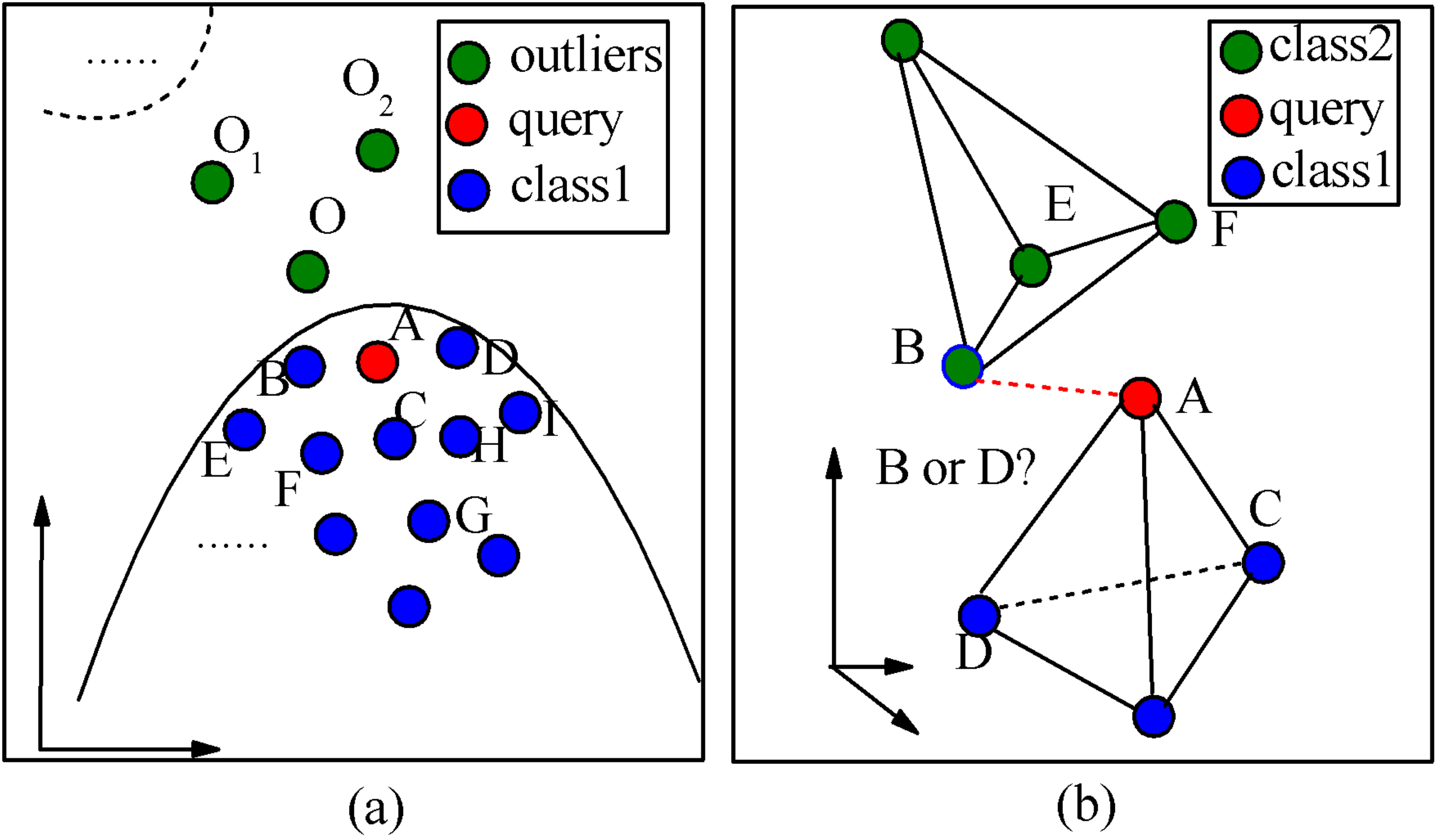}}
\caption{Examples of images that are closer to the query image
but have lower correlation with the query image. (a) Outliers are located
near the query.(b) The margin of different manifold are close to each other.}
\label{fig1}
\end{figure}

Fig1(a) illustrates the influence of outliers on the candidate image set
which is constructed by Jaccard coefficient. Given a query image $A \in M$ and
image set $\{B,C,D,E,F,G,H,I,O,O_1 ,O_2 \}$, where $M$ is the manifold which
query image $A$ located on, and $B,C,D,E,F,G,H,I \in M$. The Jaccard
coefficient of two images are presented by $J_{ij}$, and its weight is
$w_{ij} $\footnote{ The relationship of two images $i,j$ proportional to
$w_{ij} $, where $w_{ij} = w_{ji} $. $w_{ij} $ can be computed by Jaccard
coefficient.}, where $i,j \in \Omega $, $\Omega = \{A,B,C,D,O\}$. Denote the
neighborhood of image $i$ as $N_i $ which is constructed by KNN method.
Then, $N_o = \{A,B,O_1 ,O_2 \}$, $N_A = \{O,B,C,D\}$, $N_B = \{A,O,E,F\}$,
$N_C = \{A,F,H,G\}$, $N_D = \{A,C,H,I\}$, $J_{AO} = 3 \mathord{\left/
{\vphantom {3 7}} \right. \kern-\nulldelimiterspace} 7$, $J_{AB} = 3
\mathord{\left/ {\vphantom {3 7}} \right. \kern-\nulldelimiterspace} 7$,
$J_{AC} = 2 \mathord{\left/ {\vphantom {2 8}} \right.
\kern-\nulldelimiterspace} 8$, $J_{AD} = 3 \mathord{\left/ {\vphantom {3 7}}
\right. \kern-\nulldelimiterspace} 7$. The candidate image set $\Omega $ of
query $A$ is constructed by Jaccard coefficient, and is composed of the
returned top 5 images. Nevertheless, the result is unreasonable. For the
reason that $O$ is not located on $M$, which means to say $O$ is not correlated to
$A$. Moreover, since $C \in M$ and $O \notin M$, $J_{AC} < J_{AO} $ does not
meet the constraint that the candidate image set should locate on the same
manifold with $A$. In Fig1(b), we define two manifolds $M_1 $ ,$M_2$, where
$A,C,D \in M_1 $, $B,E,F \in M_2 $, and assume $w_{AC} > w_{AB} > w_{AD}
$\footnote{ If $w_{AB} = w_{AD} $, this case relates to uncertainty learning \cite{li2007artificial}. This uncertain case can be solved by our method.} and $w_{DC} + w_{DA} > w_{BC} + w_{BA}
$. If the manifold ranking methods of transductive are not taken into
consideration and only the distance weights for re-ranking is considered, the
retrieval result of $A$ is $A \leftarrow C \leftarrow B$. Clearly, for $A,C
\in M_1 $, $C$ is a reasonable retrieval result. However, $B$ is
an unreasonable result because of $B \in M_2$.

In this paper, we proposed a novel approach for image unsupervised
re-ranking of single feature on graph and illustrated rationality of multi-graph
fusion ranking by using probability theory. Our image retrieval process can be brief described as follows: (1)KNN of query $\rightarrow$ (2) CKNNS construction by TTNG $\rightarrow$ (3) Multi-feature fusion ranking by MFR.  On single feature re-ranking stage,
treble tiers neighborhood graph is proposed to improve the ranking list and offer weights of CKNNS. A multi-feature re-ranking method is designed by
the weights of improved ranking list. Then, we consider all the images in
CKNNS to re-rank the final ranking list, which is different from graph
fusion ranking in reference \cite{Zhang2015Query}\cite{Bai2016Sparse}. The main contributions of this paper
are both on single feature re-ranking and multi-graph ranking stages of
images, which are list as follows:

(1) A novel structure (TTNG) for single feature re-ranking is robust to
outlier of CKNNS in most cases.

(2) Our robust fusion re-ranking method considering CKNNS can enhance the retrieval
results with undesirable data distribution.

(3) We utilize probability theory to illustrate that the retrieval results
may be improved while more independent features involved.

(4) Our method is inductive and efficient for new sample, which need not re-construct neighborhood graph for out-of-sample extensions comparing with transductive ranking method (eg. manifold ranking or multi-graph ranking \cite{Zhao2014Affective} etc.)

\section{Related Work}

Single image feature can only describe an image from a certain view. Such
as: HSV \cite{Deselaers2008Features}, which can only get the global color information of image,
Convolutional Neural Network (CNN) \cite{Krizhevsky2012ImageNet} can extract biological features of the image, and the Bag of words
(BOW) \cite{Csurka2004Visual} uses SIFT-based \cite{David1999Object} to get the local information and the parts
distribution of the image.

In recent years, multi-feature fusion ranking have drawn lots of attentions in information retrieval field. Researches show that graph structure can be extended to
multiple features, such as hyper-graph \cite{Huang2010Image} and the manifold graph ranking
method \cite{Zhao2014Affective}. However, the MR-based methods can only get transductive ranking
and low efficiency. In order to solve the above problems, Xu et al.
proposed an efficient MR \cite{Xu2011Efficient}, which uses clustering techniques to find the
landmark and to realize inductive ranking of image datasets. Combined with
the above analysis, most of the current MR-based methods require calculation of the KNN for all images, which is a time consuming work. In order to realize
efficient inductive ranking, Zhang et al. \cite{Zhang2015Query} proposed graph density (GD) re-ranking method
based on Jaccard similarity \cite{Levandowsky1971Distance}. This method only needs to calculate the
CKNNS of query, and employs the Jaccard similarity as a judgment basis of nearest
neighbor to enhance the single feature image retrieval accuracy. It is worth
noting that the proposed method can be easily carried out by multi feature
fusion ranking and obtains a more satisfactory result. Bai et al. \cite{Bai2016Sparse} claim
the neighbors in CKNNS contribute equally, which is not a reasonable
approach in Jaccard-based method. Sparse Contextual Activation (SCA) is
proposed to improve Jaccard-based ranking by Gaussian kernel distance and
enhance performance by local consistency enhancement (LCE).

\section{Three Tiers Neighborhood Graph}

To enhance the performance in image retrieval, re-ranking is a feasible
approach for both supervised and unsupervised methods. As a supervised way,
relevance feedback (RF) selects positive/negative samples in CKNNS to
construct feedback model. And this procedure should be done for several
times. The selection is a non-automatic process in RF. Consequently,
unsupervised approaches have attracted attention in recent researches. Zhang
et al. proposed graph construction by determining Jaccard coefficient of
CKNNS and its neighborhood, which is considered in our graph construction
method (TTNG). For simplicity, we substitute ``sub-graph'' for ``graph'' in this paper.


Denote a collection of image set $X = \left\{ {x_1 , \cdots ,x_n }
\right\}$, $x^q$ is the query (or center sample, CS), $N_k \left( {x^q}
\right) = \left\{ {x_c^q } \right\}$ is the CKNNS of $x^q$, $c = 1, \cdots
,k$. CKNNS of $x^q$ is consisted of the $k$ nearest neighbors of $x^q$ under
a certain similarity measure and is also the original ranking list which
returns top-$k$ images of $x^q$.

\subsection{A Brief Review of Single Tier Neighborhood Graph (STNG)}

In this subsection, we give a brief review of graph construction re-ranking
which only needs the first single tier in reference \cite{Zhang2015Query}.

The basic idea of the re-ranking method is reciprocal neighbor relation for
two different CKNNS of samples which always indicates visual similarity of
images \cite{Zhang2015Query}\cite{Bai2016Sparse} and relationship of users in social networks \cite{Jiang2013Understanding}. Jaccard
coefficient is employed to measure the similarity of $x \in X$ and $x^q$ as
follows:

\begin{equation}
\label{eq1}
J\left( {x,x^q} \right) = \frac{\left| {N_{k_2 } \left( x \right) \cap
N_{k_1 } \left( {x^q} \right)} \right|}{\left| {N_{k_2 } \left( x \right)
\cup N_{k_1 } \left( {x^q} \right)} \right|}
\end{equation}

The weight of $x$ and $x^q$ is defined by $w\left( {x,x^q} \right) = \alpha
J\left( {x,x^q} \right)$, where $\alpha $ is a decay coefficient. The new
re-ranking list of $x$ is according to the descendant sorting of $w$. More
details can be referred in reference \cite{Zhang2015Query}. Bai et al. claim that CKNNS of
the query contributes equally in the above process and utilize Gaussian
kernel to improve this short coming. However, this approach is not always
suitable, which details in reference \cite{Bai2016Sparse}. How to choose a distance function
(only using L1-norm or L2-norm etc.) relies on data distribution. In this
paper, we only use Jaccard coefficient, and propose a novel structure in
re-ranking process in next subsection.

\subsection{TTNG}

TTNG uses Jaccard coefficient to build a weight graph as the first tier.
Then based on the first tier, we build the second tier. And we build the
third tier based on the second tier. The process of TTNG is described in Fig. 2. TTNG utilizes the neighbors of query on the same manifold to delete the outliers, and solves the problem which is detailed in Fig. 1(a).

\begin{figure*}[htbp]
\centerline{\includegraphics[width=6.5in]{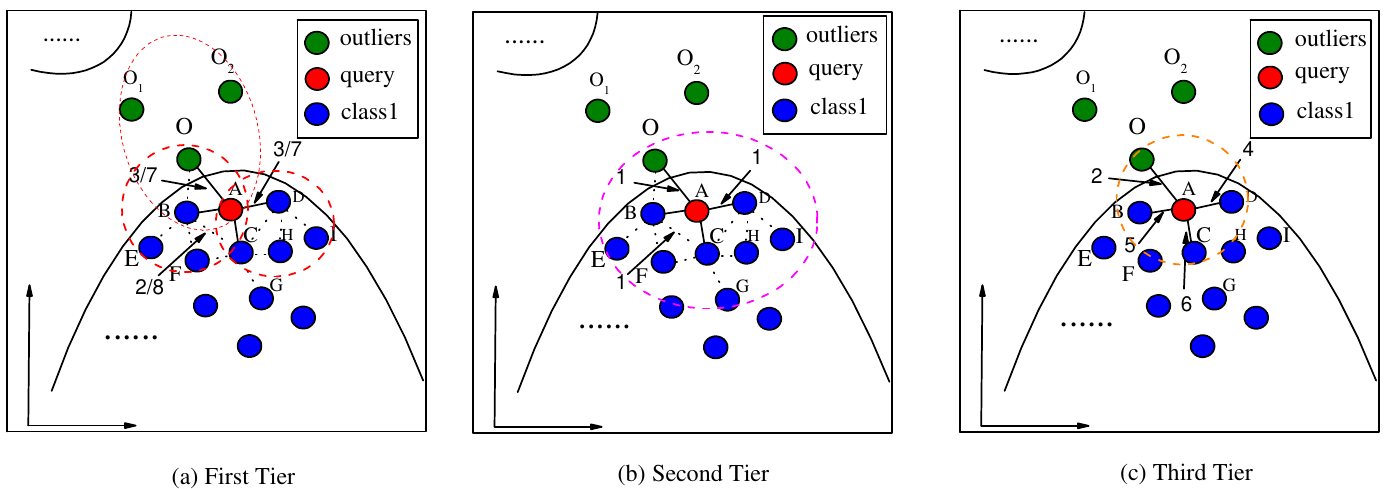}}
\caption{ The process of TTNG graph construction}
\label{fig2}
\end{figure*}

As described in STNG, we build the first tier. This motivates us to get the
weight of $x$ and $x^q$ by computing the weight ${w}'\left( {x,x^q} \right)$
of $x$ and $x^q$ as follows£º

\begin{equation}
\label{eq2}
{w}'\left( {x,x^q} \right) = \left\{ {{\begin{array}{*{20}c}
 1 \hfill & {\left( {J\left( {x,x^q} \right) > 0} \right) \wedge \left( {x
\in N_{k_1 } \left( {x^q} \right)} \right)} \hfill \\
 0 \hfill & {else} \hfill \\
\end{array} }} \right.
\end{equation}

Then, we treat ${w}'\left( {x,x^q} \right)$ as the weight of the second
tier. The weight of $x$ and $x^q$ can be computed by Eq.(\ref{eq3}) as
follows£º

\begin{equation}
\label{eq3}
w\left( {x,x^q} \right) = \sum\limits_{{x}' \in N_{k_2 } \left( {x}
\right)} {{w}'\left( {{x}',x} \right)}
\end{equation}

Finally, $w\left( {x,x^q} \right)$ is regarded as the weight of the third
tier.

From the above Eq. (\ref{eq2}, \ref{eq3}), we can see that the value of $w\left(
{x,x^q} \right)$ is determined by the Jaccard similarity coefficient of
nodes in the second and the third tiers. $w\left( {x,x^q} \right)$ in our
method is more reliable than that in STNG. To prove our method, we give
mathematical expectation of $w\left( {x,x^q} \right)$ as follows:

\begin{equation}
\label{eq4}
\begin{array}{l}
E\left( {w\left( {x,x^q} \right)} \right) = E\left( {\sum\limits_{{x}' \in
N_{k_1 } \left( {x} \right)} {{w}'\left( {{x}',x} \right)} } \right)\\
{\kern 55pt} =
E\left( {\sum\limits_{{x}' \in X} {\left[ {J\left( {{x}',x} \right) > 0}
\right]} } \right)\\
 \end{array}
\end{equation}

To simply describe the correlation of images,   we involve indicator function
$C\left( { \cdot , \cdot } \right)$, where $C\left( {x_i ,x_j } \right) = 1$
indicates the correlation of $x_i $ and $x_j $ for $\forall x_i, x_j \in X$, otherwise $C\left( {x_i ,x_j
} \right) = 0$. We also illustrate the relationship between $w\left( {x,x^q}
\right)$ and $P\left( {C\left( {x,x^q} \right) = 1} \right)$ by the following
assumption: (1) If $J\left( {x,x^q} \right)$ is not zero, $x$ and $x^q$ are
correlative, which means $C\left( {x,x^q} \right) = 1$, otherwise $x$ and
$x^q$ are not correlative, which means $C\left( {x,x^q} \right) = 0$. (2)
Images in $N_k \left( {x^q} \right)$ contribute equally to $x^q$.

By equation (\ref{eq4}), we can express $E\left( {w\left( {x,x^q} \right)} \right)$
as follows£º

\begin{equation}
\label{eq5}
\begin{array}{l}
 E\left( {w\left( {x,x^q} \right)} \right) = \sum\limits_{{x}' \in N_k
\left( {x^q} \right)} {E\left( {[J\left( {{x}',x} \right) > 0]}
\right)} \\
 {\kern 55pt} = \sum\limits_{{x}' \in N_k
\left( {x} \right)} {E\left( {C\left( {{x}',x} \right)} \right)} \\
 {\kern 55pt} = \sum\limits_{{x}' \in N_k \left( {x}
\right)} {P\left( {C\left( {{x}',x} \right) = 1} \right)} \\
 {\kern 55pt}= P\left( {C\left( {x,x^q} \right) =
1\left| {x \in N_k \left( {x} \right)} \right.} \right) \cdot k \\
 \end{array}
\end{equation}

By Eq.(\ref{eq5}), we can know that the weight function $w$ of this paper is
proportional to the probability of similar images. In this subsection, TTNG ranking list can be got by Eq.(\ref{eq3}) for image retrieval, which is corresponding to theoretical proof of Eq.(\ref{eq4}, \ref{eq5}).

\textbf{Remark}: \textit{TTNG can not deal with the problem in Fig 1 (b). This because boundary samples with high relationship (computing by STNG) on adjacent manifolds will not be distinguished by TTNG.}

\section{Multi-graph Fusion Ranking}

Fusion ranking is an essential approach for multi-feature image retrieval.
In this section, we are going to detail MFR method and the basic of MFR
theory.

\subsection{Graph Fusion}

In order to obtain the complementary information of image features to
improve the accuracy of image retrieval, we need to fuse several features of
images. We denote $V$ as node, $E$ as edge and $w$ as weight in image graph.
Assuming $m$ features have been extracted from an image. Then $m$ graphs can
be constructed by TTNG. In graph fusion methods, the $j$-th
feature graph is defined as $G^j = \left( {V^j,E^j,w^j} \right)$, where $j =
1,2, \cdots ,m$, Multi-feature graph can be expressed by $G = \left( {V,E,w}
\right)$ which satisfies three constrains as follows: 1) $V =
\bigcup\nolimits_{j = 1}^m {V^j} $; 2) $E = \bigcup\nolimits_{j = 1}^m {E^j}
$; 3) $w\left( {\hat {x},x^q} \right) = \sum\nolimits_{j = 1}^m {w^j\left(
{\hat {x},x^q} \right)} $. The fusion process is shown in Fig. 3. From
Eq.(\ref{eq5}), if the image features are independent in corresponding graph
$G^j = \left( {V^j,E^j,w^j} \right)$, according to The Law of Large
Numbers and Eq.(\ref{eq5}), fusion weight $w\left( {x,x^q} \right)$
satisfies:

\begin{equation}
\label{eq6}
P\left( {\left| {w\left( {x,x^q} \right) \cdot m - S_p} \right| > \varepsilon } \right)\buildrel
{m \to \infty } \over \longrightarrow 0
\end{equation}

where $S_p=\sum\limits_{1 \le j \le
m} {P\left( {\left. {C\left( {x,x^q} \right) = 1} \right|x \in N_k^j \left(
{x^q} \right)} \right) \cdot k}$. Eq.(\ref{eq6}) reflects that, when more independent features are involved, the
retrieval results will be better by using this method. By graph construction
and fusion method in this paper, we can construct the edge weight function
$w$ which is linear to the probability that a pair of images are similar. In
next subsection, we will introduce MFR working process for multi-feature
image retrieval.


\subsection{Re-ranking by Multi-feature}

We define $k$ images in final retrieval list of $x^q$ as $U_k \left(
{x^q} \right) = \{u_1^q ,u_2^q , \cdots ,u_k^q \}$, where $u_1^q = x^q$, $k
= 2,3, \cdots $. If the $k + 1$-\textit{th} image $u_{k + 1}^q $ needs to be added to
ranking list, previous method\cite{Bai2016Sparse} \cite{Zhang2015Query} only considers $w\left( {u_{k + 1}^q
,x^q} \right)$ which is computed by using Eq.(\ref{eq1}) , while ours compares the
weights from $w\left( {u_1^q ,x^q} \right)$ to $w\left( {u_k^q ,x^q}
\right)$. In MFR, $u_{k + 1}^q $ is selected by computing
the maximum probability $P_{k + 1} = P\left( {\left. {C\left( {i,x^q}
\right) = 1} \right|U_k \left( {x^q} \right)} \right)$ as follows:

\begin{equation}
\label{eq7}
\begin{array}{l}
u_{k + 1}^q = \mathop {\arg \max }\limits_i \left( {P\left( {\left. {C\left(
{i,x^q} \right) = 1} \right|U_k \left( {x^q} \right)} \right)} \right) \\ {\kern 22pt}=
\mathop {\arg \max }\limits_i \left( {\prod\limits_{u \in U_k \left( q
\right)} {P\left( {C\left( {u,i} \right) = 1} \right)} } \right)\\
\end {array}
\end{equation}

To avoid zero solution in Eq.(\ref{eq7}), we modify Eq.(\ref{eq7}) to the following
expression:

\begin{equation}
\label{eq8}
u_{k + 1}^q = \mathop {\arg \max }\limits_i \left( {\sum\limits_{u \in U_k
\left( q \right)} {P\left( {C\left( {u,i} \right) = 1} \right)} } \right)
\end{equation}

The final re-ranking results of MFR can be obtained by Eq.(\ref{eq8}).


\textbf{Remark}: In our re-ranking method, we assume that \textit{query and the nearest neighbor of query are embedded on the same manifold}. If this is not satisfied, the
nearest neighbor of query has been deleted by TTNG before (Fig 1(a)).

\section{Experimental results and analysis}

This section discribes the datasets and features which are used in the
experiments in details, and then analyzes the results on each dataset.

\subsection{Datasets}

Four benchmark datasets are utilized to evaluate our ranking method as follows:
UK-bench, Corel-1K, Corel-10K, and Cifar-10.Parameters of Datasets (Image Size (IS), Number of Categories (NC), Number of Each Categories (NEC), Total Images (TI)) are detailed in Table 1\footnote{In our experiments, we set $k_1 = k_2 = k$.}.

\begin{center}
\begin{table}[htbp]
\caption{Attributes of experimental dataset}
\begin{tabular}
{cccccc}
\hline
Database&
IS&
NC&
NEC&
TI&
$k$ \\
\hline
Ukbench&
640$\times $480&
2550&
4&
10.2k&
5 \\

Corel-1K&
384$\times $256&
10&
100&
1k&
50 \\

Corel-10K&
Vary&
100&
100&
10k&
50 \\

Cifar-10&
32$\times $32&
10&
6000&
60k&
50 \\
\hline
\end{tabular}
\label{tab1}
\end{table}
\end{center}

\subsection{Image features}

The main features used in this paper include: CNN, HSV and SIFT-based, etc..

HSV: HSV color space is popular for image descriptor. For one image, RGB to
HSV is a nonlinear transformation. The HSV color space is more suitable for
human perception than RGB. HSV color histogram uses 20$\times $10$\times
$10 bins for H, S and V components, respectively.

SIFT-based: Each image uses VLFeat-library to extract dense SIFT features of
images. SIFT is used to construct BOW and VOC image features by 300 (one
image with 1200 dimensional vector generated by tow layers Spatial Pyramid) and 1M
vocabulary, respectively.

CNN: We use AlexNet \cite{Krizhevsky2012ImageNet} based on convolution neural network structure, and
pre-train in the imagenet-1000 dataset for CNN. Finally, we use the L5
value as the CNN feature in image retrieval.

\subsection{UK-bench dataset}

To illustrate the effectiveness of this method in multi-feature graph
fusion, this paper combines the single feature methods which are described
in this subsection. We compare the retrieval performance of our method with eight competitive methods, including SCC\cite{Takahashi2015Mixture}, MF\cite{wang2012unsupervised}, LGD\cite{Iakovidou2015Localizing}, GD\cite{Zhang2015Query}, SR\cite{Yang2015Submodular}, CBE\cite{Zheng2014Coupled}, SCA\cite{Bai2016Sparse}, and AFF\cite{Zhou2015Augmented}. The fusion results are shown in Table 2.

\begin{center}
\begin{table}[htbp]
\caption{The performance of our method by fusion ranking in Ukbench}
\begin{tabular}
{cp{0.3cm}p{0.3cm}p{0.3cm}p{0.005cm}p{0.3cm}p{0.3cm}p{0.3cm}p{0.3cm}}
\hline
Index&
\multicolumn{3}{c}{TTNG} &
&
\multicolumn{4}{c}{Ours (TTNG-MFR)}  \\
\cline{1-4}
\cline{6-9}
SIFT-based&
$\surd $&
&
&
&
$\surd $&
&
$\surd $&
$\surd $ \\

HSV&
&
$\surd $&
&
&
$\surd $&
$\surd $&
&
$\surd $ \\

CNN&
&
&
$\surd $&
&
&
$\surd $&
$\surd $&
$\surd $ \\
\cline{1-4}
\cline{6-9}
NS-Score&
3.71&
3.52&
3.51&
&
3.88&
3.81&
3.89&
3.91 \\
\hline
\end{tabular}
\label{tab3}
\end{table}
\end{center}

\begin{center}
\begin{table}[htbp]
\caption{The N-S score of different methods in Ukbench dataset}
\begin{tabular}
{p{0.5cm}p{0.5cm}p{0.5cm}p{0.5cm}p{0.5cm}p{0.5cm}p{0.5cm}p{0.5cm}p{0.5cm}}
\hline
SCC&
MF&
LGD&
GD&
SR&
CBE&
SCA&
AFF&
Ours \\
\hline
3.66&
3.68&
3.76&
3.77&
3.78&
3.79&
3.86&
3.88&
\textbf{3.91} \\
\hline
\end{tabular}
\label{tab4}
\end{table}
\end{center}

\begin{center}
\begin{figure}[htbp]
\centerline{\includegraphics[width=3.5 in]{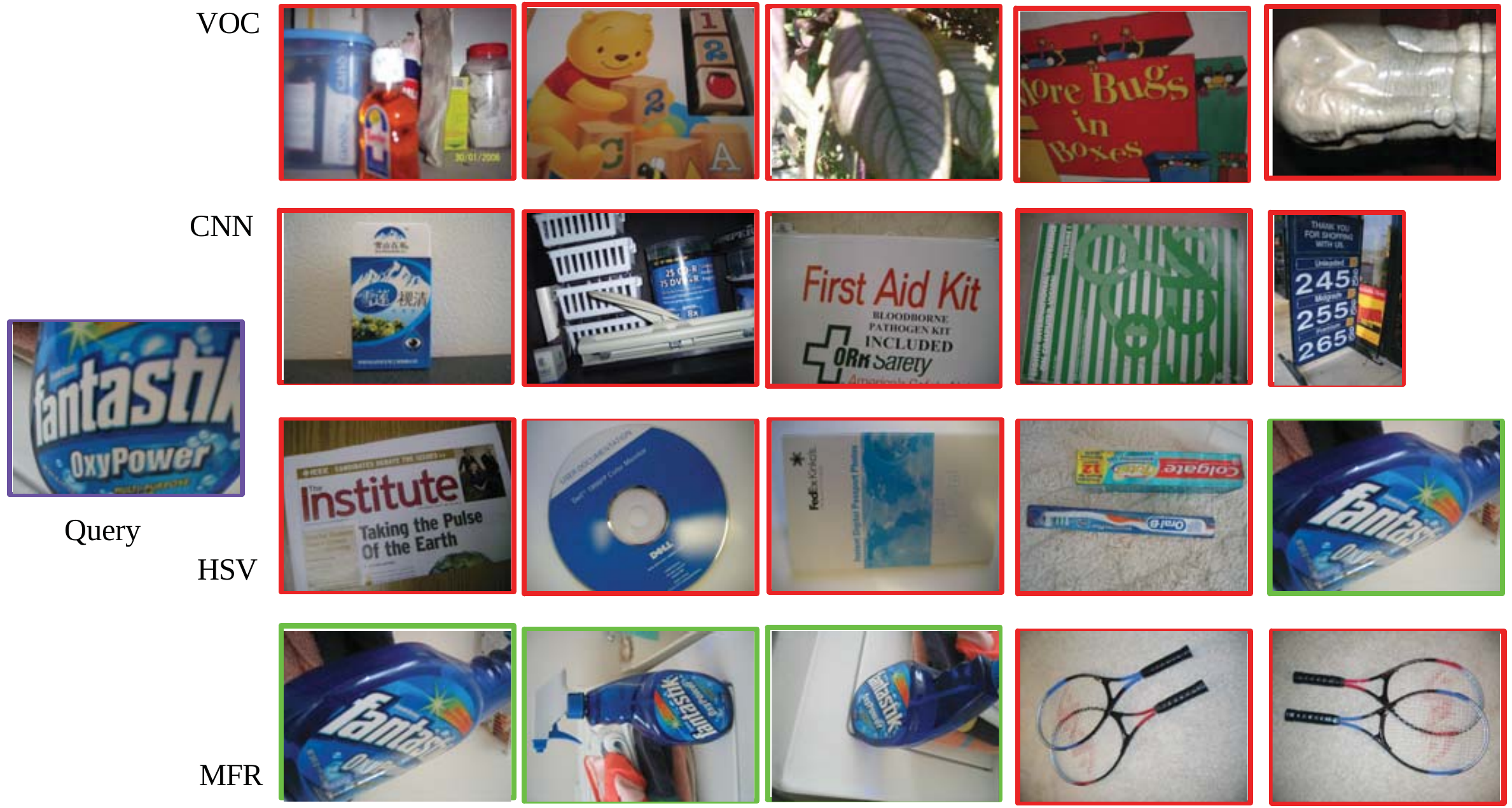}}
\caption{The final ranking list for a query example in Ukbench dataset by
TTNG-MFR}
\label{fig4}
\end{figure}
\end{center}

The results of our method and other state-of-the-art methods in the Ukbench data
set are shown in Table 3. Our method achieve N-S score 3.91 in the Ukbench dataset by
using global color feature HSV, local features SIFT-based, ans deep featur CNN,
which is shown in Table 3. This illustrates that the
more independent features are involved, the better fusion results are
achieved (N-Sscore=3.91). To illustrate above point of view, a visual retrieval
example is conducted as follows: we use an image as query in Ukbench dataset
(see Fig. 3). The fusion ranking results of multi-feature which fuse VOC, CNN, and HSV (the last row in Table 2) are better than that of each of
single feature (the first three rows in Table 2).

\subsection{Corel-1K and Corel-10K datasets}

In order to show the effects of TTNG-MFR on different datasets, this
subsection conducts experiments on the Corel-1K and Corel-10K datasets. We
compare the precisions of the first 20 returning images in Corel-1K dataset. We involve seven state-of-the-art image retrieval method, including BFF \cite{Guo2013Image}, ECF \cite{Walia2014Fusion}, SCQ \cite{Zeng2015Image}, PCM \cite{Yu2011Colour}, CTF \cite{Lin2009A}, and GD \cite{Zhang2015Query}. In Corel-10K data
set, we compare our method with SSH \cite{Liu2015Content}, Ri-HOG \cite{Chen2015Content}, HOG \cite{Chen2015Content}, and GD \cite{Zhang2015Query} to  evaluate our method\footnote{We use HSV, SIFT-based, and CNN features in Corel-10K data
set.}.

\begin{center}
\begin{table}[htbp]
\caption{The performance of our method in Corel-1K with 20 returns by single feature({\%})}
\begin{tabular}
{cccccc}
\hline
\raisebox{-1.50ex}[0cm][0cm]{HSV}&
\raisebox{-1.50ex}[0cm][0cm]{BOW}&
\raisebox{-1.50ex}[0cm][0cm]{CNN}&
\multicolumn{3}{c}{method}  \\
\cline{4-6}
 &
 &
 &
origin&
TTNG&
Ours \\
\hline
$\surd $&
&
&
71.67&
72.44&
\textbf{74.93} \\

&
$\surd $&
&
67.96&
68.03&
\textbf{70.22} \\
&
&
$\surd $&
57.97&
\textbf{58.80} &
57.09 \\
\hline
\end{tabular}
\label{tab4}
\end{table}
\end{center}

\begin{center}
\begin{table}[htbp]
\caption{The performance of our method by fusion ranking in Corel-1K ({\%})}
\begin{tabular}
{ccccc}
\hline
\raisebox{-1.50ex}[0cm][0cm]{BOW}&
\raisebox{-1.50ex}[0cm][0cm]{HSV}&
\raisebox{-1.50ex}[0cm][0cm]{CNN}&
\multicolumn{2}{c}{Ours}  \\
\cline{4-5}
 &
 &
 &
12-precision&
20-precision \\
\hline
$\surd $&
&
$\surd $&
89.73&
89.01 \\

&
$\surd $&
$\surd $&
76.48&
72.41 \\

$\surd $&
$\surd $&
&
79.94&
76.70 \\

$\surd $&
$\surd $&
$\surd $&
\textbf{92.43}&
\textbf{90.52} \\
\hline
\end{tabular}
\label{tab5}
\end{table}
\end{center}

\begin{center}
\begin{table*}[htbp]
\caption{The precision of previous works and our method with 20 returns in
Corel-1K dataset({\%})}
\begin{tabular}
{cccccccccccc}
\hline
\raisebox{-1.50ex}[0cm][0cm]{Methods}&
\multicolumn{11}{c}{Classes}  \\
\cline{2-12}
 &
African&
Beach&
Building&
Bus&
Dinosaur&
Elephant&
Flower&
Horse&
Mountains&
Food&
Avg \\
\hline
BFF&
84.70&
45.4&
67.80&
85.30&
99.30&
71.10&
93.30&
95.80&
49.80&
80.80&
77.30 \\

ECF&
51.00&
\textbf{90.00}&
58.00&
78.00&
78.00&
\textbf{100.00}&
84.00&
\textbf{100.0}&
84.00&
38.00&
78.30 \\

SCQ&
72.50&
65.20&
70.60&
\textbf{89.20}&
\textbf{100.00}&
70.50&
94.80&
91.80&
72.25&
78.80&
80.57 \\

PCM&
84.9&
35.6&
61.6&
81.8&
\textbf{100.00}&
59.1&
93.1&
92.8&
40.4&
68.2&
71.7 \\

CTF&
68.3&
54.0&
56.2&
88.8&
99.3&
65.8&
89.1&
80.3&
52.2&
73.3&
72.7 \\

GD&
83.75&
64.65&
67.85&
69.55&
94.15&
83.75&
78.45&
93.30&
85.65&
81.05&
80.22 \\

\textbf{Ours}&
\textbf{95.25}&
72.2&
\textbf{82.2}&
83.55&
\textbf{100.00}&
95.2&
\textbf{99.05}&
99.05&
\textbf{91.85}&
\textbf{86.8}&
\textbf{90.52} \\
\hline
\end{tabular}
\label{tab6}
\end{table*}
\end{center}

\begin{center}
\begin{table}[htbp]
\caption{The precision and recall of different image descriptors and
ranking methods with 12 returns in Corel-10K dataset ({\%})}
\begin{tabular}
{cp{0.7cm}cp{0.6cm}p{0.6cm}p{0.6cm}}
\hline
Type $\backslash$ method&
SSH&
Ri-HOG&
HOG&
GD&
Ours \\
\hline
Precision&
54.88&
53.13 &
33.29&
60.75&
\textbf{65.00} \\
Recall&
6.59&
6.25&
3.94&
7.29&
\textbf{7.80 }\\
\hline
\end{tabular}
\label{tab7}
\end{table}
\end{center}

\begin{center}
\begin{table*}[htbp]
\caption{Performance of different methods in Cifar-10 dataset ({\%})}
\begin{tabular}
{cccccccc}
\hline
Method&
BOW&
HSV&
CNN&
12-precision&
20-precision&
50-precision&
100-precision \\
\hline
Origin&
$\surd $&
&
&
47.52&
43.62&
38.39&
35.21 \\

Origin&
&
$\surd $&
&
30.84&
27.07&
22.85&
20.78 \\

Origin&
&
&
$\surd $&
52.91&
49.27&
44.11&
40.68 \\

STNG &
$\surd $&
&
&
47.52&
43.63&
38.39&
35.21 \\

STNG &
&
$\surd $&
&
31.11&
27.12&
22.52&
20.61 \\

STNG &
&
&
$\surd $&
52.93&
49.27&
44.07&
40.64\\

GD&
$\surd $&
&
$\surd $&
52.56&
48.25&
41.15&
36.84 \\

Ours&
$\surd $&
&
$\surd $&
\textbf{55.33}&
\textbf{52.04}&
\textbf{47.21}&
\textbf{44.22} \\
\hline
\end{tabular}
\label{tab8}
\end{table*}
\end{center}

The ranking results for different features are shown in Table 4. As
seen in Table 4, the re-ranking precisions of TTNG-MFR can be improved by
HSV feature (+3.26{\%}) and is decreased by CNN feature (-0.88{\%}). The different features
of the TTNG structure is important to Re-ranking results. Therefore, the higher the
accuracy of the original feature is, the better the retrieval performance of the
Re-ranking method is.

For multi-feature image retrieval task, the experimental results of TTNG-MFR
method are shown in Table 5. As seen from Table 5, the precision of
the top 20 images in corel-1K is 90.52{\%} which is the highest precision among
fusion CNN, BOW and HSV features in Table 5. Results of our method are consistent with those
of UK-bench dataset.

By comparing the results of Table 4 and Table 5, we can see that our method
is suitable for fusion of independent features. Furthermore, In order to
illustrate the effectiveness of TTNG-MFR, we compare TTNG-MFR with other
methods in corel-10K and corel-1K datasets and give the experimental
results in Table 6 and Table 7, respectively.

\subsection{Cifar-10 dataset}

We conduct experiments on the Cifar-10 dataset, and compare precisions of
the top 10, 20, 50 and 100 retrieved images to illustrate effectiveness of TTNG-MFR
for large-scale image retrieval.

The experimental results from Table 8 show that the single feature
re-ranking improves performance little on cifar-10 dataset. This result
shows that the re-ranking method does not work well on cifar-10 dataset.
By fusing single features in our method, the precision of the top 100 images on Cifar-10
dataset is increased by 3.54{\%}, while that of the graph density method is
decreased by 3.84{\%} after fusion ranking, which illustrates that MFR method of
this paper is effective and robust to the fusion of independent features.
Therefore, TTNG-MFR can obtain high precision and is efficiency for large scale
image retrieval task.

\subsection{Time Complexity }

Time complexity of our method on-line correlates with $k$ and $m$, and
especially has no correlation with the size of dataset. The values of $k$ and
$m$ are selected based on actual situation. The ranking result of images'
single feature in dataset can be calculated off-line. For each image,
Re-ranking time complexity is $O\left( 2 \cdot k \cdot log_2(k)\right)$, time complexity of Jaccard
coefficient is $O\left( {\left( {m + 2} \right) \cdot k^2 \cdot \log _2
\left( k \right)} \right)$, and time complexity of multi-feature fusion is
$O\left( {\left( {m + 1} \right) \cdot k} \right)$. Thus, the extra time
complexity generated by our method is $O\left( {\left( {m + 2} \right) \cdot k^2 \cdot \log _2
\left( k \right)} + {\left( {m + 1} \right) \cdot k} + 2\cdot k\cdot log_2(k)\right)$, which is detailed in Table
9 (Intel(R) Core(TM) i7-4770 CPU @ 3.40GHz 3.40GHz).

\begin{table}[htbp]
\caption{The Re-rank Time (in ms) on the Test Datasets}
\begin{tabular}
{cccc}
\hline
Dataset&
Ours(ms)& $m$ & $k$ \\
\hline
Ukbench&
0.01 & 3&6\\
Corel-1k&
0.50 & 3& 20\\
Corel-10k & 1.52 & 3& 25\\
 Cifar-10&
 2.66 & 2& 50\\
\hline
\end{tabular}
\label{tab9}
\end{table}

\section{Conclusion}

In this paper, we propose two important techniques and demonstrate essential
of graph fusion in multi-feature image retrieval task. TTNG can remove the
impact of the outlier who is near the retrieval image, which shows the
robustness of our method. This strategy can get better results than STNG.
Furthermore, MFR can remove the impact of the other classes. As shown in
TTNG, although the operation to sum the weights of different features is
simple, it is adequately efficient. We evaluate our TTNG-MFR on four
benchmark datasets, which demonstrates the effectiveness and efficiency of
our ranking method.

In general, fusion re-ranking requires that low-dimensional manifold of each
feature maintains structure of original image manifold. If a feature
extraction technology destroy structure of original image manifold, better performance can not be obtain by our method. For our
further work, we are going to put feature extraction and fusion re-ranking
into the same scenario.


%
\bibliographystyle{aaai}
\bibliography{Ref}
\end{document}